\documentclass[10pt,journal]{IEEEtran}
\usepackage{graphicx}
\usepackage{amsmath,amssymb,times,cite}
\usepackage{booktabs}

\usepackage{color}
\newcommand{\op}[1]{\operatorname{#1}}
\newcommand{\bg}[1]{\boldsymbol{#1}} 
\newcommand{\bm}[1]{\mathbf{#1}} 
\newcommand{\vc}[3]{\overset{#2}{\underset{#3}{#1}}}

\newcommand\T{{\mathpalette\raiseT\intercal}}
\newcommand\raiseT[2]{%
\setbox0\hbox{$#1{#2}$}\raise\dp0\box0}

\title{\LARGE\textbf{Regular Splitting Graph Network for 3D Human Pose Estimation}}
\author{Md. Tanvir Hassan and A. Ben Hamza\\
Concordia Institute for Information Systems Engineering\\
Concordia University, Montreal, QC, Canada
}
\date{}

\begin{document}
\maketitle

\begin{abstract}
In human pose estimation methods based on graph convolutional architectures, the human skeleton is usually modeled as an undirected graph whose nodes are body joints and edges are connections between neighboring joints. However, most of these methods tend to focus on learning relationships between body joints of the skeleton using first-order neighbors, ignoring higher-order neighbors and hence limiting their ability to exploit relationships between distant joints. In this paper, we introduce a higher-order regular splitting graph network (RS-Net) for 2D-to-3D human pose estimation using matrix splitting in conjunction with weight and adjacency modulation. The core idea is to capture long-range dependencies between body joints using multi-hop neighborhoods and also to learn different modulation vectors for different body joints as well as a modulation matrix added to the adjacency matrix associated to the skeleton. This learnable modulation matrix helps adjust the graph structure by adding extra graph edges in an effort to learn additional connections between body joints. Instead of using a shared weight matrix for all neighboring body joints, the proposed RS-Net model applies weight unsharing before aggregating the feature vectors associated to the joints in order to capture the different relations between them. Experiments and ablations studies performed on two benchmark datasets demonstrate the effectiveness of our model, achieving superior performance over recent state-of-the-art methods for 3D human pose estimation.
\end{abstract}


\section{Introduction}
The objective of 3D human pose estimation is to predict the positions of a person's joints in still images or videos. It is one of the most rapidly evolving computer vision technologies, with diverse real-world applications ranging from activity recognition and pedestrian behavior analysis~\cite{YZhao:20} to sports and safety surveillance in assisted living retirement homes. In healthcare, for instance, potential benefits of human pose estimation include posture correction during exercise and rehabilitation of the limbs, thereby helping people adopt a healthy lifestyle.
	
Existing 3D human pose estimation methods can be broadly categorized into two main streams: single-stage~\cite{li20143d} and two-stage approaches~\cite{pavlakos2017coarse,sun2017compositional}. Single-stage methods typically use a deep neural network to regress 3D keypoints from images in an end-to-end manner. On the other hand, two-stage approaches, also referred to as lifting methods, consist of two decoupled stages. In the first stage, 2D keypoints are extracted from an image using an off-the-shelf 2D pose detector such as the cascaded pyramid network~\cite{chen2018cascaded} or the high-resolution network~\cite{sun2019deep}. In the second stage, the extracted 2D keypoints are fed into a regression model to predict 3D poses~\cite{martinez2017simple,yang20183d,fang2018learning,rayat2018exploiting,pavlakos2018ordinal,sharma2019monocular}. These keypoints include the shoulders, knees, ankles, wrists, pelvis, hips, head, and others on the human skeleton. Two-stage approaches generally outperform the single-stage methods thanks, in part, to recent advances in 2D pose detectors, particularly the high-resolution representation learning networks that learn not only semantically strong representations, but are also spatially precise~\cite{sun2019deep}. For example, Martinez \textit{et al.}~\cite{martinez2017simple} introduce a simple two-stage approach to 3D human pose estimation by designing a multilayer neural network with two blocks comprised of batch normalization, dropout, and a rectified linear unit activation function. This multilayer network also uses residual connections to facilitate model training and improve generalization performance.
	
Recently, graph convolutional networks (GCNs) and their variants have emerged as powerful methods for 2D-to-3D human pose estimation~\cite{zhao2019semantic,zou2020high,quan2021higher,liu2020comprehensive,zou2021modulated,Guoliang:22} due largely to the fact that a 2D human skeleton can naturally be represented as a graph whose nodes are body joints and edges are connections between neighboring joints. For example, Zhao \textit{et al.}~\cite{zhao2019semantic} propose a semantic GCN architecture to capture local and global node relationships that are learned through end-to-end training, resulting in improved 3D pose estimation performance. While graph neural networks, particularly GCNs, have shown great promise in effectively tackling the 3D human pose estimation problem, they suffer, however, from a number of issues. First, GCNs focus primarily on learning relationships between body joints using first-order neighbors, ignoring higher-order neighbors; thereby limiting their ability to exploit relationships between distant joints. This challenge can be mitigated using higher-order graph neural networks~\cite{abu2019mixhop}, which have proven effective at capturing long-range dependencies between body joints~\cite{zou2020high,quan2021higher}. Second, GCNs share the transformation matrix in the graph convolutional filter for all nodes, hindering the efficiency of information exchange between nodes, especially for a multi-layer network. To overcome this limitation, Liu \emph{et al.} \cite{liu2020comprehensive} introduce various weight unsharing mechanisms and apply different feature transformations to graph nodes before aggregating the associated features. The downside of these mechanisms is that they increase the model size by a factor equal of the number of body joints. This challenge can be alleviated by incorporating both weight and affinity modulation into the shared weight matrix and adjacency matrix, respectively~\cite{zou2021modulated} in order to help improve model generalization.

Another recent line of work leverages Transformer architectures, which employ a multi-head self-attention mechanism, to capture spatial and temporal information from 2D pose sequences~\cite{PoseFormer:2021,li2022mhformer}. While Transformer-based architectures are able to encode long-range dependencies between body joints in the spatial and temporal domains, they often require large-scale training datasets to achieve comparable performance in comparison with their convolutional networks counterparts, particularly on visual recognition tasks. This can make training and inference computationally expensive. Moreover, the attention mechanism used in Transformers involves computing an attention score between every pair of tokens in the input sequence, which can be computationally expensive, especially for longer sequences. More recently, Zhuang \emph{et al.}~\cite{liu2022convnet} have proposed ConvNeXt architecture, competing favorably with Transformers in terms of accuracy and scalability, while maintaining the simplicity and efficiency of standard convolutional networks. Similar to the Transformer block and unlike the ResNet block, the ConvNeXt block is comprised of convolutional layers, followed by layer normalization and a Gaussian error linear unit activation function~\cite{liu2022convnet}.
	
To address the aforementioned issues, we introduce a higher-order regular splitting graph network, dubbed RS-Net, for 3D human pose estimation by leveraging regular matrix splitting together with weight and adjacency modulation. The layer-wise propagation rule of the proposed method is inspired by the iterative solution of a sparse linear system via regular splitting. We follow the two-stage approach for 3D human pose estimation by first applying a state-of-art 2D pose detector to obtain 2D pose predictions, followed by a lifting network for predicting the 3D pose locations from the 2D predictions. The key contributions of this work can be summarized as follows:
	
\begin{itemize}
\item We propose a higher-order regular splitting graph network for 3D human pose estimation using matrix splitting in conjunction with weight and adjacency modulation.
\item We introduce a new objective function for training our proposed graph network by leveraging the regularizer of the elastic net regression.
\item We design a variant of the ConvNeXT residual block and integrate it into our graph network architecture.
\item We demonstrate through experiments and ablation studies that our proposed model achieves state-of-the-art performance in comparison with strong baselines.
\end{itemize}

\medskip\noindent The rest of this paper is structured as follows. In Section II, we review related work in the area of 3D pose estimation. In Section III, we summarize the basic notation and concepts. In Section IV, we formulate the learning task at hand and then describe the main building blocks of the proposed graph network architecture, including a generalization to higher-order settings. In Section V, we present empirical results comparing our model with state-of-the-art approaches for 3D pose estimation on two large-scale standard benchmarks. Finally, we conclude in Section VI by summarizing our key contributions and pointing out future work directions.

\section{Related Work}
Both graph convolutional networks and 3D human pose estimation have received a flurry of research activity over the past few years. Here, we only review the techniques most closely related to ours. Like much previous work discussed next, we approach the problem of 3D human pose estimation using a two-stage pipeline.

\medskip\noindent\textbf{Graph Convolution Networks.}\quad GCNs and their variants have recently become the method of choice in graph representation learning, achieving state-of-the-art performance in numerous downstream tasks~\cite{defferrard2016convolutional,Kipf:17,Velickovic:17,chen2020simple}, including 3D human pose estimation~\cite{zhao2019semantic,zou2021modulated,liu2020comprehensive,Ailing2021Hard}. However, GCNs apply graph convolutions in the one-hop neighborhood of each node, and hence fail to capture long-range relationships between body joints. This weakness can be mitigated using higher-order graph convolution filters~\cite{abu2019mixhop} and concatenating the features of body joints from multi-hop neighborhoods with the aim of improving model performance in 3D human pose estimation~\cite{zou2020high,quan2021higher}. To capture higher-order information in the graph, Wu \textit{et al.}~\cite{FWu:19} also propose a simple graph convolution by removing the nonlinear activation functions between the layers of GCNs and collapsing the resulting function into a single linear transformation using the normalized adjacency matrix powers.

\medskip\noindent\textbf{Transformer and MLP-based Architectures.}\quad Transformer-based models have shown promising results in 3D human pose estimation and are an active area of research~\cite{Weixi2021GraFormer,PoseFormer:2021,Jinlu2022Mix,Zhang2022Aware,Kevin2021End,Sha2022PSTMO,Li2022Strided,Einfalt2023Uplift}. A Transformer encodes 2D joint positions into a series of feature vectors using a self-attention mechanism, which allows the model to capture long-range dependencies between different joints and to attend to the most relevant joints for predicting the 3D joint positions. For example, Zheng \textit{et al.}~\cite{PoseFormer:2021} introduce PoseFormer, a spatio-temporal approach for 3D human pose estimation in videos that combines spatial and temporal transformers. This approach uses two separate transformers, one for modeling spatial information and the other for modeling temporal information. The spatial transformer focuses on modeling the 2D spatial relationships between the joints of the human body, while the temporal transformer models the temporal dependencies between frames. However, Poseformer only estimates human poses from the central frame of a video, which may not provide sufficient context for accurate pose estimation in complex scenarios. While Transformers have shown great potential in 3D human pose estimation, they typically require large amounts of labeled data to train effectively and are designed to process sequential data. Also, as with any spatio-temporal method, the quality of the input video can significantly impact the accuracy of the model's pose estimations. In contrast, GCNs are specifically designed for processing graph-structured data, more efficient on sparse data, produce interpretable feature representations, and require less training data to achieve good performance.

Motivated by the good performance of the MLP-mixer model~\cite{tolstikhin2021mlp} in image classification tasks, Li \textit{et al.}~\cite{Wenhao22GraphMLP} propose GraphMLP, a neural network architecture comprised of multilayer perceptrons (MLPs) and GCNs, showing competitive performance in 3D human pose estimation. GraphMLP integrates the graph structure of the human body into an MLP model, which facilitates both local and global spatial interactions. It employs a GCN block to aggregate local information between neighboring joints and a prediction head to estimate the 3D joint positions.

\medskip\noindent\textbf{3D Human Pose Estimation.}\quad The basic goal of 3D human pose estimation is to predict the locations of a human body joints in images or videos. To achieve this goal, various methods have been proposed, which can learn to categorize human poses. Most of these methods can be classified into one-stage approaches that regress 3D keypoints from images using deep neural networks in an end-to-end manner~\cite{li20143d} or two-stage approaches that employ an off-the-shelf 2D pose detector to extract 2D keypoints and then feed them into a regression model to predict 3D poses~\cite{yang20183d,fang2018learning,rayat2018exploiting,pavlakos2018ordinal,sharma2019monocular,ge20193d,pavllo20193d,zhao2019semantic,YujunCai:19,HaiCi:2019,
liu2020comprehensive,zou2020high}.

Our proposed graph neural network falls under the category of 2D-to-3D lifting. While GCNs have proven powerful at learning discriminative node representations on graph-structured data, they usually extract first-order neighborhood patterns for each joint, ignoring higher-order neighborhood information and hence limiting their ability to exploit relationships between distant joints. Moreover, GCNs share the same feature transformation for each node, hampering the efficiency of information exchange between body joints. Our work differs from existing approaches in that we use higher-order neighborhoods in combination with weight and adjacency modulation in order to not only capture long-range dependencies between body joints, but also learn additional connections between body joints by adjusting the graph structure via a learnable modulation matrix. In addition, we design a variant of the ConvNeXt block and integrate it into our model architecture with the goal of improving accuracy in human pose estimation, while maintaining the simplicity and efficiency of standard convolutional networks.
	
\section{Preliminaries}
\noindent\textbf{Basic Notions.}\quad Consider a graph $\mathcal{G}=(\mathcal{V},\mathcal{E})$, where $\mathcal{V}=\{1,\ldots,N\}$ is the set of $N$ nodes and $\mathcal{E}\subseteq \mathcal{V}\times\mathcal{V}$ is the set of edges. In human pose estimation, nodes correspond to body joints and edges represent connections between two body joints. We denote by $\bm{A}=(a_{ij})$ an $N\times N$ adjacency matrix (binary or real-valued) whose $(i,j)$-th entry $a_{ij}$ is equal to the weight of the edge between neighboring nodes $i$ and $j$, and 0 otherwise. Two neighboring nodes $i$ and $j$ are denoted as $i\sim j$, indicating that they are connected by an edge. We denote by $\mathcal{N}_i =\{j\in\mathcal{V}: i\sim j\}$ the neighborhood of node $i$. We also denote by $\bm{X}=(\bm{x}_{1},...,\bm{x}_{N})^{\T}$ an $N\times F$ feature matrix of node attributes, where $\bm{x}_{i}$ is an $F$-dimensional row vector for node $i$.
	
\medskip\noindent\textbf{Spectral Graph Theory.}\quad The normalized Laplacian matrix is defined as
\begin{equation}
\bm{L}=\bm{I}-\bm{D}^{-1/2}\bm{A}\bm{D}^{-1/2}=\bm{I}-\hat{\bm{A}},
\end{equation}
where $\bm{D}=\op{diag}(\bm{A}\bm{1})$ is the diagonal degree matrix, $\bm{1}$ is an $N$-dimensional vector of all ones, and $\hat{\bm{A}}=\bm{D}^{-1/2}\bm{A}\bm{D}^{-1/2}$ is the normalized adjacency matrix. Since the normalized Laplacian matrix is symmetric positive semi-definite, it admits an eigendecomposition given by $\bm{L}=\bm{U}\bg{\Lambda}\bm{U}^{\T}$, where $\bm{U}=(\bm{u}_1,\dots,\bm{u}_N)$ is an orthonormal matrix whose columns constitute an orthonormal basis of eigenvectors and $\bg{\Lambda}=\op{diag}(\lambda_1,\dots,\lambda_N)$ is a diagonal matrix comprised of the corresponding eigenvalues such that $0=\lambda_1\le\dots\le\lambda_N\le 2$ in increasing order~\cite{Chung:97}. Hence, the eigenvalues of the normalized adjacency matrix lie in the interval $[-1,1]$, indicating that the spectral radius (i.e., the highest absolute value of all eigenvalues) $\rho(\hat{\bm{A}})$ is less than 1

\medskip\noindent\textbf{Regular Matrix Splitting.}\quad Let $\bm{S}$ be an $N\times N$ matrix. The decomposition $\bm{S}=\bm{B}-\bm{C}$ is called a regular splitting of $\bm{S}$ if $\bm{B}$ is nonsingular and both $\bm{B}^{-1}$ and $\bm{C}$ are nonnegative matrices~\cite{Saad:03}. Using this matrix splitting, the solution of the matrix equation $\bm{S}\bm{x}=\bm{r}$, where $\bm{r}$ is a given vector, can be obtained iteratively as follows:
\begin{equation}
\bm{x}^{(t+1)} = \bm{B}^{-1}\bm{C}\bm{x}^{(t)}+\bm{B}^{-1}\bm{r},
\end{equation}
where $\bm{x}^{(t)}$ and $\bm{x}^{(t+1)}$ are the $t$-th and $(t+1)$-th iterations of $\bm{x}$, respectively. This iterative method is convergent if and only if the spectral radius of the iteration matrix $\bm{B}^{-1}\bm{C}$ is less than 1. It can also be shown that given a regular splitting, $\rho(\bm{B}^{-1}\bm{C})<1$ if and only if $\bm{S}$ is nonsingular and its inverse is nonnegative~\cite{Saad:03}.

\section{Proposed Method}
In this section, we first start by defining the learning task at hand, including the objective function. Then, we present the main components of the proposed higher-order regular splitting graph network with weight and adjacency modulation for 3D human pose estimation.

\subsection{Problem Statement}
Let $\mathcal{D}=\left\{\left(\mathbf{x}_{i}, \mathbf{y}_{i}\right)\right\}_{i=1}^{N}$ be a training set of 2D joint positions $\mathbf{X}=$ $\left(\mathbf{x}_{1}, \ldots, \mathbf{x}_{N}\right)^{\T} \in \mathbb{R}^{N \times 2}$ and their associated 3D joint positions $\mathbf{Y}=\left(\mathbf{y}_{1}, \ldots, \mathbf{y}_{N}\right)^{\T} \in \mathbb{R}^{N \times 3} .$ The goal of 3D human pose estimation is to learn the parameters $\bm{w}$ of a regression model $f: \mathbf{X} \rightarrow \mathbf{Y}$ by finding a minimizer of the following loss function
\begin{equation}	
\bm{w}^{*}=\arg\min_{\bm{w}}\frac{1}{N}\sum_{i=1}^{N}l(f(\bm{x}_{i}),\bm{y}_{i}),
\end{equation}
where $l(f(\bm{x}_{i}),\bm{y}_{i})$ is an empirical loss function defined by the learning task. Since human pose estimation is a regression task, we define $l(f(\bm{x}_{i}),\bm{y}_{i})$ as a weighted sum (convex combination) of the $\ell_2$ and $\ell_1$ loss functions
\begin{equation}	
l(f(\bm{x}_{i}),\bm{y}_{i})=(1-\alpha)\sum_{i=1}^{N}\Vert\bm{y}_{i}-f(\bm{x}_{i})\Vert_{2}^{2}+
\alpha\sum_{i=1}^{N}\Vert\bm{y}_{i}-f(\bm{x}_{i})\Vert_{1},
\label{eq:loss}
\end{equation}
where $\alpha\in [0,1]$ is a weighting factor controlling the contribution of each term. It is worth pointing out that our proposed loss function~\eqref{eq:loss} is inspired by the regularizer used in the elastic net regression technique~\cite{HuiZou:05}, which is a hybrid of ridge regression and lasso regularization.

\subsection{Spectral Graph Filtering}
The goal of spectral graph filtering is to use polynomial or rational polynomial filters defined as functions of the graph Laplacian in order to attenuate high-frequency noise corrupting the graph signal. Since the normalized Laplacian matrix is diagonalizable, applying a spectral graph filter with transfer function $h$ on the graph signal $\bm{X}\in\mathbb{R}^{N\times F}$ yields
\begin{equation}
\bm{H}=h(\bm{L})\bm{X}=\bm{U}h(\bg{\Lambda})\bm{U}^{\T}\bm{X}=\bm{U}\op{diag}(h(\lambda_i))\bm{U}^{\T}\bm{X},
\label{Eq:SGF}
\end{equation}
where $\bm{H}$ is the filtered graph signal. However, computing all the eigenvalue/eigenvectors of the Laplacian matrix is notoriously expensive, particularly for very large graphs. To circumvent this issue, spectral graph filters are usually approximated using Chebyshev polynomials~\cite{Taubin:96,Hammond:11,defferrard2016convolutional} or rational polynomials~\cite{Levie:18,Bianchi:22,Wijesinghe:19}.
	
\subsection{Implicit Fairing Filter}
The implicit fairing filter is an infinite impulse response filter whose transfer function is given by $h_{s}(\lambda)=1/(1+s\lambda)$, where $s$ is a positive parameter~\cite{Desbrun:99,quan2021higher}. Substituting $h$ with $h_s$ in Eq.~\eqref{Eq:SGF}, we obtain
\begin{equation}
\bm{H}=(\bm{I}+s\bm{L})^{-1}\bm{X},
\end{equation}
where $\bm{I}+s\bm{L}$ is a symmetric positive definite matrix (all its eigenvalue are positive), and hence admits an inverse. Therefore, performing graph filtering with implicit fairing is equivalent to solving the following sparse linear system:
\begin{equation}
(\bm{I}+s\bm{L})\bm{H}=\bm{X},
\label{Eq:IF}
\end{equation}
which can be efficiently solved using iterative methods~\cite{Saad:03}.	

\subsection{Regular Splitting and Iterative Solution}
\noindent\textbf{Regular Splitting.}\quad For notational simplicity, we denote $\bm{L}_s = \bm{I}+s\bm{L}$, which we refer to as the implicit fairing matrix. Using regular splitting, we can split the matrix $\bm{L}_s$ as follows:
\begin{equation}
\bm{L}_s=(1+s)\bm{I}-s\hat{\bm{A}}=\bm{B}-\bm{C},
\label{Eq:RegSplit}
\end{equation}
where
$$\bm{B}=(1+s)\bm{I} \quad\text{and}\quad \bm{C}=s\hat{\bm{A}}=s\bm{D}^{-1/2}\bm{A}\bm{D}^{-1/2}.$$
Note that $\bm{B}$ is a scaled identity matrix and $\bm{C}$ is a scaled normalized adjacency matrix. It should be noted that for both matrices, the scaling is uniform (i.e., constant scaling factors). Since $\hat{\bm{A}}$ is a nonnegative matrix and its spectral radius is less than 1, it follows that $\rho(s\hat{\bm{A}})<s+1$. Therefore, the implicit fairing matrix $\bm{L}_s$ is an $M$-matrix, and consequently its inverse is a nonnegative matrix. In words, an $M$-matrix can be defined as a matrix with positive diagonal elements, nonpositive off-diagonal elements and a nonnegative inverse.

\medskip\noindent\textbf{Iterative Solution.}\quad Using regular splitting, the implicit fairing equation \eqref{Eq:IF} can be solved iteratively as follows:
\begin{equation}
\begin{split}
\bm{H}^{(t+1)}
&=\bm{B}^{-1}\bm{C}\bm{H}^{(t+1)} + \bm{B}^{-1}\bm{X}\\
&=(s/(1+s))\hat{\bm{A}}\bm{H}^{(t)}+(1/(1+s))\bm{X},
\end{split}
\label{Eq:ISJ}
\end{equation}
Since the spectral radius of the normalized adjacency matrix $\hat{\bm{A}}$ is smaller than 1, it follows that the spectral radius of the iteration matrix $\bm{B}^{-1}\bm{C}$ is less than $s/(1+s)$, which is in turn smaller than 1. Therefore, the iterative method is convergent. This convergence property can also be demonstrated by noting that $\bm{L}_{s}$ is nonsingular and its inverse is nonnegative; thereby $\bm{B}^{-1}\bm{C}<1$.

\medskip\noindent We can rewrite the iterative solution given by Eq.~\eqref{Eq:ISJ} in matrix form as follows:
\begin{equation}
\bm{H}^{(t+1)}=\hat{\bm{A}}\bm{H}^{(t)}\bm{W}_{s}+\bm{X}\widetilde{\bm{W}}_{s},
\label{Eq:Iterative}
\end{equation}
where $\bm{W}_{s}=\op{diag}(s/(1+s))$ and $\widetilde{\bm{W}}_{s}=\op{diag}(1/(1+s))$ are $F\times F$ diagonal matrices, each of which has equal diagonal entries, and $\bm{H}^{(t)}$ is the $t$-th iteration of $\bm{H}$.

\medskip\noindent\textbf{Theoretical Properties.}\quad In the regular splitting $\bm{L}_s=\bm{B}-\bm{C}$ given by Eq.~\eqref{Eq:RegSplit}, both $\bm{L}_s$ and $\bm{B}$ are nonsingular because $\bm{L}_s$ is a symmetric positive definite matrix and $\bm{B}$ is a scaled identity matrix. Hence, the following properties hold:
\begin{itemize}
\item The matrices $\bm{B}^{-1}\bm{C}$ and $\bm{L}_{s}^{-1}\bm{C}$ commute, i.e., $\bm{B}^{-1}\bm{C}\bm{L}_{s}^{-1}=\bm{L}_{s}^{-1}\bm{C}\bm{B}^{-1}$.
\item The matrices $\bm{B}^{-1}\bm{C}$ and $\bm{L}_{s}^{-1}\bm{C}$ have the same eigenvectors.
\item If $\mu_i$ and $\tau_i$ are the eigenvalues of $\bm{B}^{-1}\bm{C}$ and $\bm{L}_{s}^{-1}\bm{C}$, respectively, then $\mu_i=\tau_{i}/(1+\tau_{i})$.
\item The regular splitting is convergent if and only if $\tau_i>-1/2$ for all $i=1,\dots,N$.
\item Since both $\bm{B}^{-1}\bm{C}$ and $\bm{L}_{s}^{-1}\bm{C}$ are nonnegative matrices, the regular splitting is convergent and
$$\rho(\bm{B}^{-1}\bm{C})=\frac{\rho(\bm{L}_{s}^{-1}\bm{C})}{1+\rho(\bm{L}_{s}^{-1}\bm{C})}.$$
\end{itemize}
Detailed proofs of these properties for a regular splitting of any matrix can be found in~\cite{Woznicki:01}.	

\subsection{Regular Splitting Graph Network}
In order to learn new feature representations for the input feature matrix of node attributes over multiple layers, we draw inspiration from the iterative solution given by Eq.~\eqref{Eq:Iterative} to define a multi-layer graph convolutional network with skip connections as follows:
\begin{equation}
\bm{H}^{(\ell+1)}=\sigma(\hat{\bm{A}}\bm{H}^{(\ell)}\bm{W}^{(\ell)} +\bm{X}\widetilde{\bm{W}}^{(\ell)}),\,\, \ell=0,\dots,L-1
\label{Eq:IS}
\end{equation}
where $\bm{W}^{(\ell)}\in\mathbb{R}^{F_{\ell}\times F_{\ell+1}}$ and $\widetilde{\bm{W}}^{(\ell)}\in\mathbb{R}^{F\times F_{\ell+1}}$ are learnable weight matrices, $\sigma(\cdot)$ is an element-wise nonlinear activation function such as the Gaussian Error Linear Unit (GELU), $\bm{H}^{(\ell)}\in\mathbb{R}^{N\times F_{\ell}}$ is the input feature matrix of the $\ell$-th layer and $\bm{H}^{(\ell+1)}\in\mathbb{R}^{N\times F_{\ell+1}}$ is the output feature matrix. The input of the first layer is the initial feature matrix $\bm{H}^{(0)}=\bm{X}$. Notice that the key difference between \eqref{Eq:Iterative} and \eqref{Eq:IS} is that the latter defines a representation updating rule for propagating node features layer-wise using trainable weight matrices for learning an efficient representation of the graph, followed by an activation function to introduce non-linearity into the network in a bid to enhance its expressive power. This propagation rule is essentially comprised of feature propagation and feature transformation. The skip connections used in the proposed model allow information from the initial feature matrix to bypass the current layer and be directly added to the output of the current layer. This helps preserve important information that may be lost during the aggregation process, thereby improving the flow of information through the network.
	
The $i$-th row of the output feature matrix can be expressed as follows:
\begin{equation}
\bm{h}_{i}^{(\ell+1)}=\sigma\Biggl(\sum_{j\in\mathcal{N}_{i}} \hat{a}_{ij}\bm{h}_{j}^{(\ell)}\bm{W^{(\ell)}} +\bm{x}_{i}\bm{\widetilde{W}^{(\ell)}} \Biggr),
\end{equation}
where $\hat{a}_{ij}$ is the $(i,j)$-th entry of the normalized adjacency matrix $\hat{\bm{A}}$ and $\bm{h}_{j}^{(\ell)}$ is the neighboring feature vector of node $i$ in the input feature matrix $\bm{H}^{(\ell)}$. In words, the feature vector of each node $i$ is updated by transforming (i.e., embedding) the feature vectors of its neighboring nodes via the same projection matrix (i.e., shared weight matrix) $\bm{W^{(\ell)}}$, followed by aggregating the transformed feature vectors using a sum aggregator and then adding them to the transformed initial feature vector. Using a shared weight matrix is, however, suboptimal for articulated body modeling due largely to the fact the relations between different body joints are different~\cite{liu2020comprehensive}. To address this limitation, Liu \textit{et al.}~\cite{liu2020comprehensive} introduce various weight unsharing mechanisms in an effort to capture the different relations between body joints, and hence improve human pose estimation performance. The basic idea is to use different weight matrices to transform the features vectors of the neighboring nodes before applying the sum aggregator:
\begin{equation}
\bm{h}_{i}^{(\ell+1)}=\sigma\Biggl(\sum_{j \in \mathcal{N}_{i}} \hat{a}_{ij}\bm{h}_{j}^{(\ell)} \bm{W}_{j}^{(\ell)} +\bm{x}_{i}\bm{\widetilde{W}}^{(\ell)}\Biggr),
\label{Eq:IT}
\end{equation}
where $\bm{W}_{j}^{(\ell)}$ is the weight matrix for feature vector $\bm{h}_{j}^{(\ell)}$ at the $\ell$-th layer. This weight unsharing mechanism is referred to as pre-aggregation because weight unsharing is applied before feature vectors' aggregation. In addition, the pre-aggregation method performs the best in 3D human pose estimation~\cite{liu2020comprehensive}.
	
\medskip\noindent\textbf{Weight Modulation.}\quad While weight unsharing has proven effective at capturing the different relations between body joints, it also increases the model size by a factor equal to the number of joints. To tackle this issue, we use weight modulation~\cite{zou2021modulated} in lieu of weight unsharing. Weight modulation employs a shared weight matrix, but learns a different modulation vector for each neighboring node $j$ according to the following update rule
\begin{equation}
\bm{h}_{i}^{(\ell+1)}=\sigma\Biggl(\sum_{j \in \mathcal{N}_{i}} \hat{a}_{i j}\bm{h}_{j}^{(\ell)}\Bigl(\bm{W}^{(\ell)}\odot\bm{m}_{j}^{(\ell)}\Bigr)  +\bm{x}_{i}\bm{\widetilde{W}}^{(\ell)}\Biggr),
\label{Eq:IU}
\end{equation}
where $\bm{m}_{j}^{(\ell)}\in\mathbb{R}^{F_{\ell+1}}$ is a learnable modulation (row) vector for each neighboring node $j$ and $\odot$ denotes element-wise multiplication.

Hence, the layer-wise propagation rule with weight modulation can be written in matrix form as follows:
\begin{equation}
\bm{H}^{(\ell+1)}=\sigma\Bigl(\hat{\bm{A}}((\bm H^{(\ell)}\bm{W^{(\ell)}})\odot\bm{M}^{(\ell)})+\bm{X}\bm{\widetilde{W}}^{(\ell)}\Bigr),
\label{Eq:IV}
\end{equation}
where $\bm{M}^{(\ell)}\in\mathbb{R}^{N\times F_{\ell+1}} $ is a weight modulation matrix whose $j$-th row is the modulation vector $\bm{m}_{j}^{(\ell)}$.

\medskip\noindent\textbf{Adjacency Modulation.}\quad Following~\cite{zou2021modulated}, we modulate the normalized adjacency matrix in order to capture not only the relationships between neighboring nodes, but also the distant nodes (e.g., arms and legs of a human skeleton)
\begin{equation}
\check{\bm{A}}= \hat{\bm{A}} + \bm{Q},
\label{Eq:IX}
\end{equation}
where $\bm{Q}\in\mathbb{R}^{N\times N}$ is a learnable adjacency modulation matrix. Since we consider undirected graphs (e.g., human skeleton graph), we symmetrize the adjacency modulation matrix $\bm{Q}$ by adding it to its transpose and dividing by 2. Therefore, the layer-wise propagation rule of the regular splitting graph network with weight and adjacency modulation is given by
\begin{equation}
\bm{H}^{(\ell+1)}=\sigma\Bigl(\check{\bm{A}}((\bm H^{(\ell)}\bm{W^{(\ell)}})\odot\bm{M}^{(\ell)})+\bm{X}\bm{\widetilde{W}}^{(\ell)}\Bigr).
\label{Eq:IW}
\end{equation}

\begin{figure*}[!htb]
\centering
\includegraphics[scale=.55]{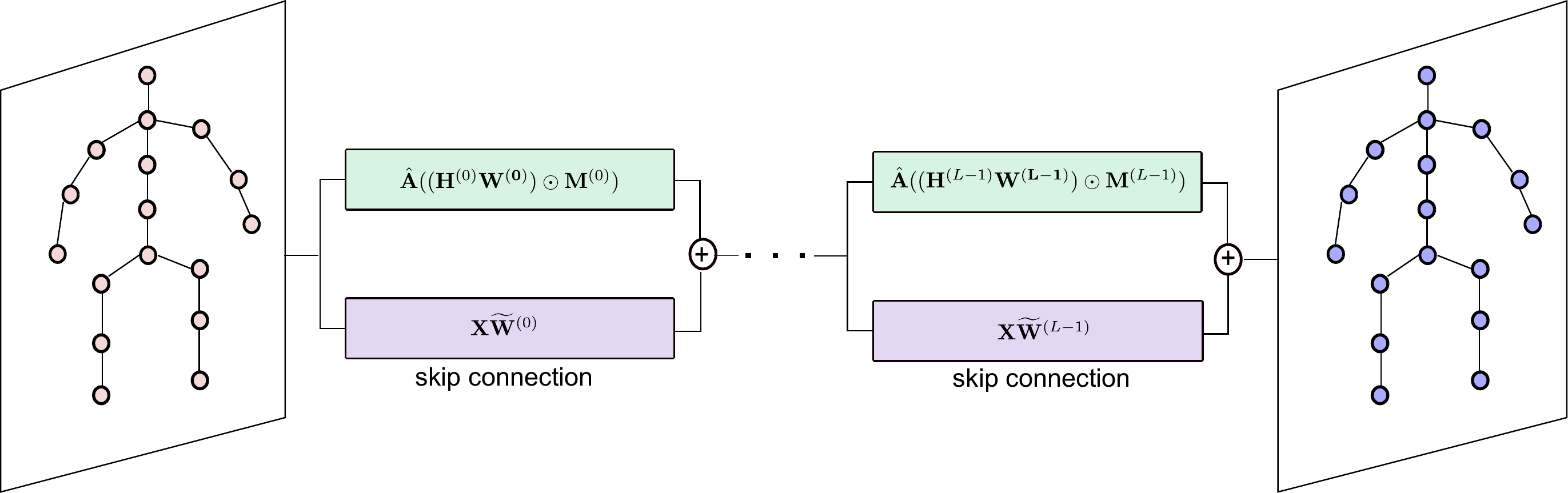}
\caption{Illustration of the layer-wise propagation rule for the proposed RS-Net model. Each block is comprised of a skip connection and a higher-order graph convolution with weight and adjacency modulation.}
\label{Fig:RSNet}
\end{figure*}

The proposed layer-wise propagation rule is illustrated in Figure~\ref{Fig:RSNet}, where each block consists of a skip connection and a higher-order graph convolution with weight and adjacency modulation. The idea of skip connection is to carry over information from the initial feature matrix. 	

\subsection{Higher-Order Regular Splitting Graph Network}
\begin{figure}[!htb]
\setlength{\tabcolsep}{1em}
\centering
\begin{tabular}{cc}
\includegraphics[scale=.26]{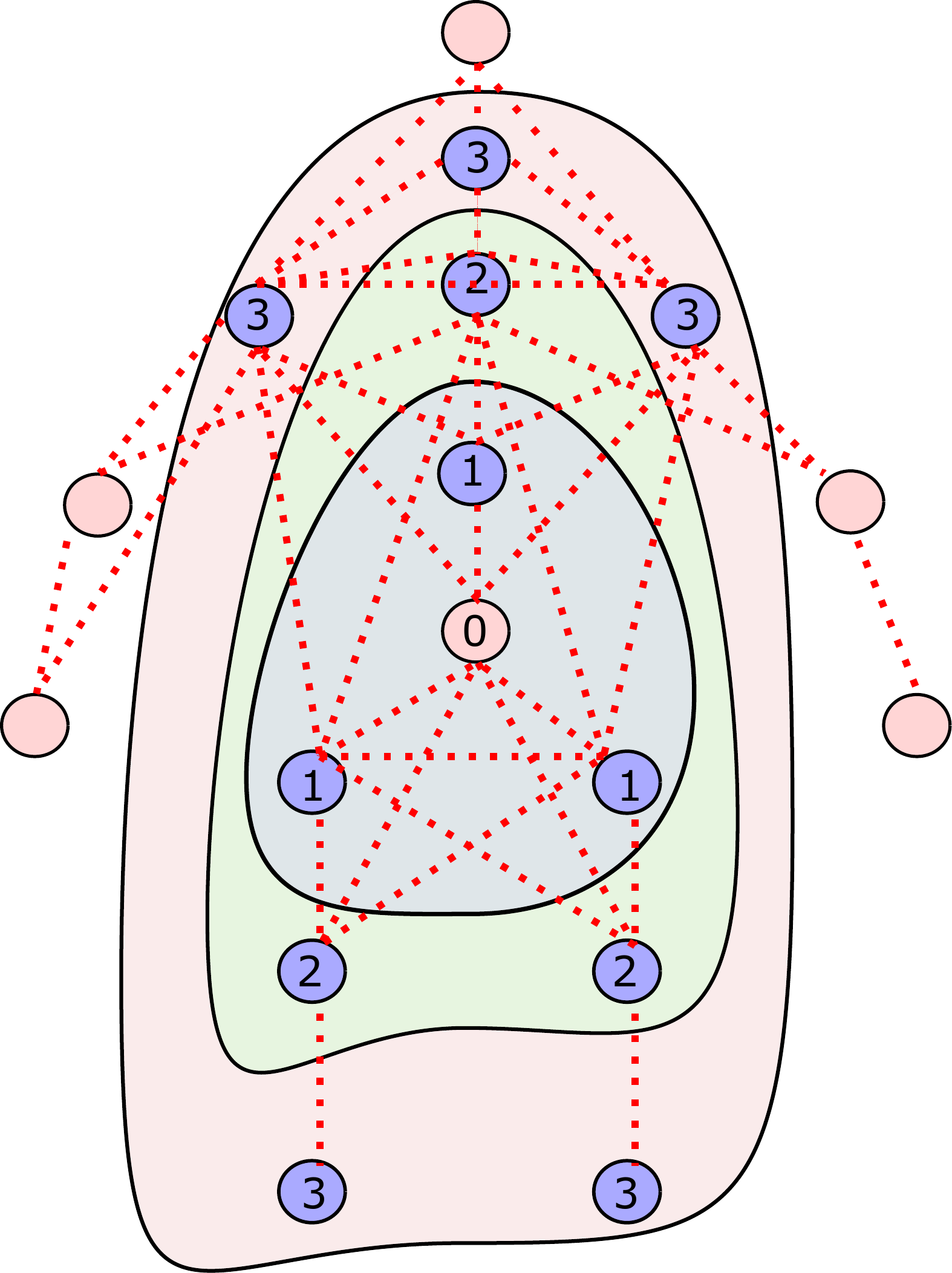} &
\includegraphics[scale=.28]{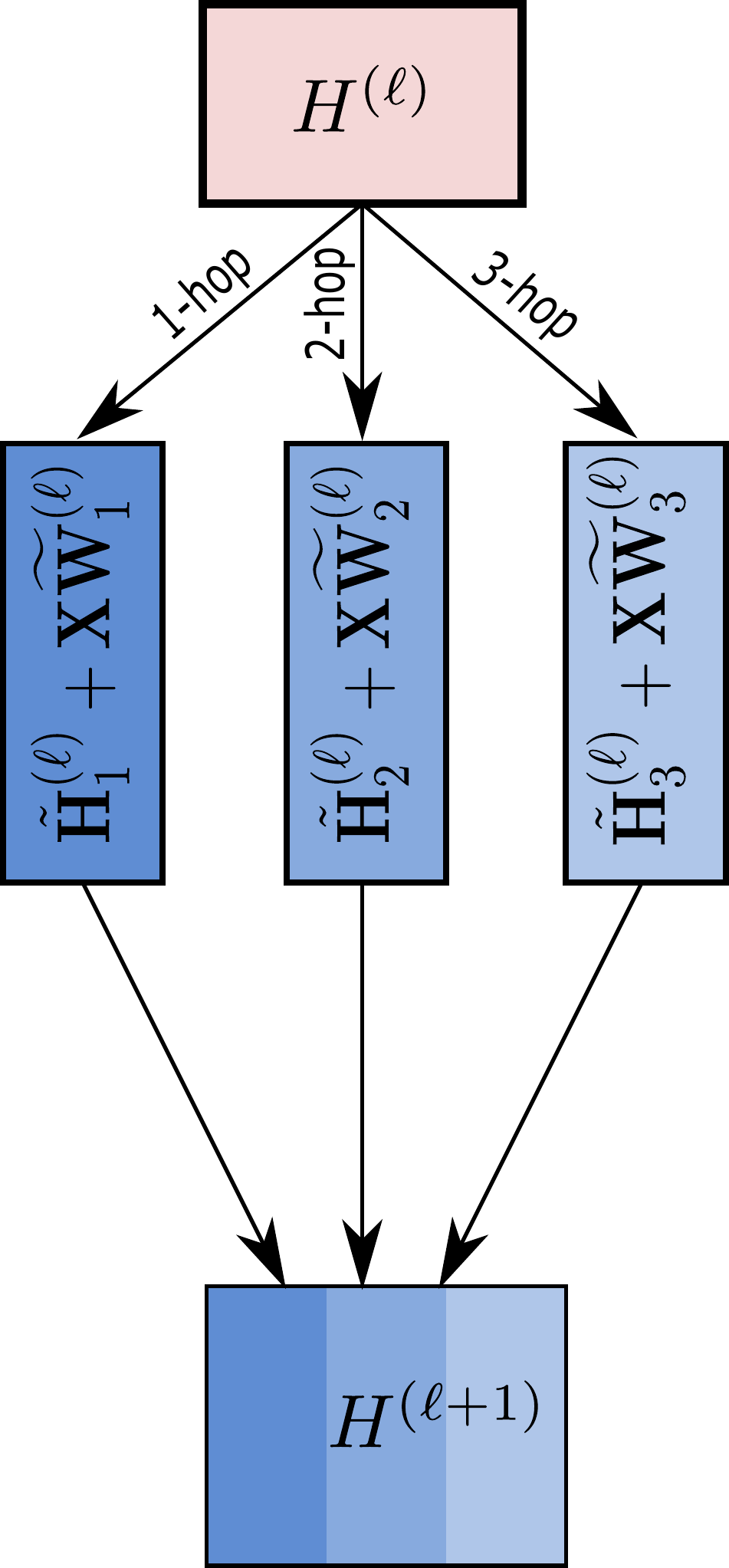}
\end{tabular}
\caption{Illustration of RS-Net feature concatenation for $K=3$ with weight and adjacency modulation. Dashed lines represent extra edges added to the human skeleton via the learnable matrix in adjacency modulation.}
\label{Fig:humangraph2}
\end{figure}

In order to capture high-order connection information and long-range dependencies, we use $k$-hop neighbors to define a higher-order regular splitting network with the following layer-wise propagation rule:
\begin{equation}
\bm{H}^{(\ell+1)}=\sigma\left(\vc{\parallel}{K}{k=1} (\tilde{\bm{H}}_{k}^{(\ell)} +\bm{X}\bm{\widetilde{W}}_{k}^{(\ell)})\right)
\end{equation}
where
\begin{equation}
\tilde{\mathbf{H}}_{k}^{(\ell)}=\check{\bm{A}}^{k}((\bm H^{(\ell)}\bm{W}_{k}^{(\ell)})\odot\bm{M}_{k}^{(\ell)})
\end{equation}
and $\check{\bm{A}}^{k}$ is the $k$-th power of the normalized adjacency matrix with adjacency modulation. The learnable weight and modulation matrices $\bm{W}_{k}^{(\ell)}$ and $\bm{M}_{k}^{(\ell)}$ are associated with the $k$-hop neighborhood, and $\parallel$ denotes concatenation. For each $k$-hop neighborhood, the node representation is updated by aggregating information from its neighboring nodes using weight and adjacency modulation, as well as carrying over information from the initial node features via skip connection. Then, high-order features are concatenated, as illustrated in Figure~\ref{Fig:humangraph2}, followed by applying a non-linear transformation. Notice how additional edges, shown as dashed lines, are created as a result of adding a learnable modulation matrix to the normalized adjacency matrix.
	
\begin{figure*}[!htb]
\centering
\includegraphics[scale=.76]{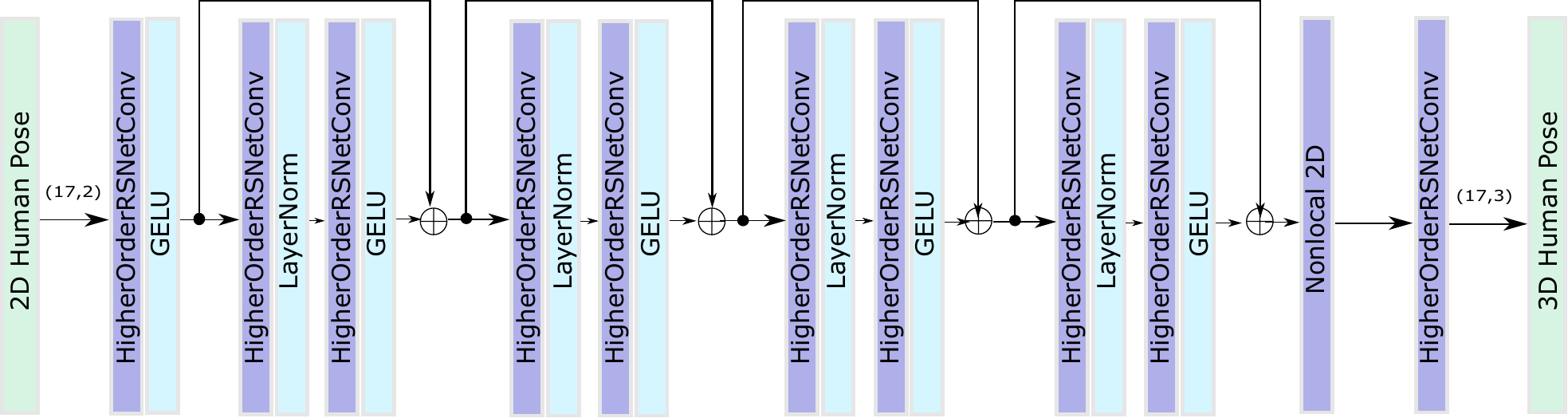}
\caption{Overview of the proposed network architecture for 3D pose estimation. Our model takes 2D pose coordinates (16 or 17 joints) as input and generates 3D pose predictions (16 or 17 joints) as output. We use ten higher-order graph convolutional layers with four residual blocks. In each residual block, the first convolutional layer is followed by layer normalization, while the second convolutional layer is followed by a GELU activation function, except for the last convolutional layer which is preceded by a non-local layer.}
\label{Fig:model}
\end{figure*}

\medskip\noindent\textbf{Model Architecture.} Figure~\ref{Fig:model} depicts the architecture of our proposed RS-Net model for 3D human pose estimation. The input consists of 2D keypoints, which are obtained via a 2D pose detector. We use higher-order regular splitting graph convolutional layers defined by the layer-wise propagation rule of RS-Net to capture long-range connections between body joints. Inspired by the architectural design of the ConvNeXt block~\cite{liu2022convnet}, we adopt a residual block comprised of two higher-order regular splitting graph convolutional (RS-NetConv) layers. The first convolutional layer followed by layer normalization, while the second convolutional layer is followed by a GELU activation function, as illustrated in Figure~\ref{Fig:model}. We also employ a non-local layer~\cite{wang2018non} before the last convolutional layer and we repeat each residual block four times.
	
\medskip\noindent\textbf{Model Prediction.}\quad The output of the last higher-order graph convolutional layer of RS-Net contains the final output node embeddings, which are given by
\begin{equation}
\hat{\bm{Y}}=(\hat{\bm{y}}_{1},\dots,\hat{\bm{y}}_{N})^{\T}\in\mathbb{R}^{N\times 3},
\end{equation}
where $\hat{\bm{y}}_{i}$ is a three-dimensional row vector of predicted 3D pose coordinates.

\medskip\noindent\textbf{Model Training.}\quad The parameters (i.e., weight matrices for different layers) of the proposed RS-Net model for 3D human pose estimation are learned by minimizing the loss function
\begin{equation}
\mathcal{L} =\frac{1}{N}\left[(1-\alpha)\sum_{i=1}^{N}\Vert\bm{y}_{i}-\hat{\bm{y}}_{i}\Vert_{2}^{2}+
\alpha\sum_{i=1}^{N}\Vert\bm{y}_{i}-\hat{\bm{y}}_{i}\Vert_{1}\right],
\end{equation}
which is a weighted sum of the mean squared and mean absolute errors between the 3D ground truth joint locations $\bm{y}_{i}$ and estimated 3D joint locations $\hat{\bm{y}}_{i}$ over a training set consisting of $N$ joints.
	
\section{Experiments}
In this section, we conduct experiments on real-world datasets to evaluate the performance of our model in comparison with competitive baselines for 3D human pose estimation. The code is available at: https://github.com/nies14/RS-Net

\subsection{Experimental Setup}
\noindent\textbf{Datasets.}\quad We evaluate our approach on two large-scale benchmark datasets: Human3.6M and MPI-INF-3DHP. Human3.6M is the most widely-used dataset in 3D human pose estimation~\cite{ionescu2013human3}, comprised of 3.6 million 3D human poses for 5 female and 6 male actors as well as their corresponding images captured from four synchronized cameras at 50 Hz. A total of 15 actions are performed by each actor in an indoor environment. These actions include directions, discussion, eating, greeting, talking on the phone, and so on. Following~\cite{zou2020high, martinez2017simple}, we apply normalization to the 2D and 3D poses before feeding the data into the model. For the MPI-INF-3DHP dataset~\cite{Dushyant:2017}, there are 8 actors performing 8 actions from 14 camera views, covering a greater diversity of poses. This dataset includes a test set of 6 subjects with confined indoor and complex outdoor scenes.
	
\medskip\noindent\textbf{Evaluation Protocols and Metrics.}\quad We adopt different metrics to evaluate the performance of our model in comparison with strong baselines for 3D human pose estimation. For the Human3.6M dataset, we employ two widely-used metrics: mean per joint position error (MPJPE) and Procrustes-aligned
mean per joint position error (PA-MPJPE). Both metrics are measured in millimeters, and lower values indicate better performance. MPJPE, also referred to as Protocol \#1, computes the average Euclidean distance between the predicted 3D joint positions and ground truth after the alignment of the root joint (central hip). PA-MPJPE, also known as Protocol \#2, is computed after rigid alignment of the prediction with respect to the ground truth. Both protocols use 5 subjects (S1, S5, S6, S7, S8) for training and 2 subjects (S9, S11) for testing. For the MPI-INF-3DHP dataset, we also employ two commonly-used evaluation metrics: Percentage of Correct Keypoints (PCK) under 150mm and the Area Under the Curve (AUC) in line with previous works~\cite{quan2021higher,zou2021compositional,yang20183d,pavlakos2018ordinal,Habibie:19,li2019generating}. Higher values of PCK and AUC indicate better performance.

\medskip\noindent\textbf{Baseline Methods.}\quad We evaluate the performance of our RS-Net model against various state-of-the-art pose estimation methods, including Semantic GCN~\cite{zhao2019semantic}, High-order GCN~\cite{zou2020high}, Weight Unsharing~\cite{liu2020comprehensive}, Compositional GCN~\cite{zou2021compositional}, and Modulated GCN~\cite{zou2021modulated}. We also compare RS-Net against Transformer-based models for 3D human pose estimation such as METRO~\cite{Kevin2021End}, GraFormer~\cite{Weixi2021GraFormer}, PoseFormer~\cite{PoseFormer:2021} and MixSTE~\cite{Jinlu2022Mix}, as well as PoseAug~\cite{Kehong2021PoseAug}, a framework for 3D human pose estimation that allows for pose augmentation through differentiable operations.
	
\medskip\noindent\textbf{Implementation Details.}\quad Following the 2D-to-3D lifting approach~\cite{li2022mhformer,YujunCai:19,pavllo20193d,zou2021modulated}, we employ the high-resolution network (HR-Net)~\cite{sun2019deep} as 2D detector and train/test our model using the detector's output. We use PyTorch to implement our model, and all experiments are conducted on a single NVIDIA GeForce RTX 3070 GPU with 8G memory. We train our model for 30 epochs using AMSGrad, a variant of ADAM optimizer, which employs the maximum of past squared gradients in lieu of the exponential average to update the parameters. For 2D pose detections, we set the batch size to 512 and the filter size to 96. We also set the initial learning rate to 0.005 and the decay factor to 0.90 per 4 epochs. The weighting factor $\alpha$ is set to 0.1. For the 2D ground truth, we set the batch size to 128 and the filter size to 64. The initial learning rate is set to 0.001 with a decay factor of 0.95 applied after each epoch and 0.5 after every 5 epochs. For $K$-hop feature concatenation, we set the value of $K$ to 3. Following~\cite{YujunCai:19}, we incorporate a non-local layer~\cite{wang2018non} and a pose refinement network to improve the performance. We also decouple self-connections from the modulated normalized adjacency matrix~\cite{liu2020comprehensive}. In addition, we apply horizontal flip augmentation~\cite{zou2021modulated, li2022mhformer}. Furthermore, to prevent overfitting we add dropout with a factor of 0.2 after each graph convolutional layer.
	
\subsection{Results and Analysis}
\noindent\textbf{Quantitative Results.}\quad In Table~\ref{Tab:Result1}, we report the performance comparison results of our RS-Net model and various state-of-the-art methods for 3D human pose estimation. As can be seen, our model yields the best performance in most of the actions and also on average under both Protocol \#1 and Protocol \#2, indicating that our RS-Net is very competitive. This is largely attributed to the fact that RS-Net can better exploit high-order connections through multi-hop neighborhoods and also learns not only different modulation vectors for different body joints, but also additional connections between the joints. Under Protocol \#1, Table~\ref{Tab:Result1} shows that RS-Net performs better than ModulatedGCN~\cite{zou2021modulated} on 13 out of 15 actions by a relative improvement of 4.86\% on average. It also performs better than high-order GCN~\cite{zou2020high} on all actions, yielding an error reduction of approximately 15.47\% on average. Moreover, our model outperforms SemGCN~\cite{zhao2019semantic} by a relative improvement of 18.40\% on average. While recent Transformer models~\cite{Weixi2021GraFormer,PoseFormer:2021,Zhang2022Aware,Jinlu2022Mix,Kevin2021End} have shown great promise in 3D human pose estimation tasks, it is important to note that most of these models are either (i) spatio-temporal methods that are specifically designed for long sequences of frames, (ii) employ dynamic graphs, or (iii) use data augmentation strategies to boost performance. Nevertheless, we compared our model against some of these strong baselines, and the results are reported in Table~\ref{Tab:Result1}. In addition, we included the results of our model using the cascaded pyramid network (CPN) as a pose detector~\cite{chen2018cascaded}, showing superior performance over the baselines for various poses and exhibits better performance on average.

\begin{table*}[!htb]
\caption{Performance comparison of our model and baseline methods using MPJPE (in millimeters) on Human3.6M under Protocol \#1. The average errors are reported in the last column. Boldface numbers indicate the best performance, whereas the underlined numbers indicate the second best performance. $(f{=}1)$ indicates that the number of frames is set to 1.}
\small
\setlength{\tabcolsep}{1pt}
\smallskip
\centering
\begin{tabular}{l*{17}{c}}
\toprule[1pt]
& \multicolumn{15}{c}{Action}\\
\cmidrule(lr){2-16}
Method & Dire. & Disc. &  Eat & Greet & Phone & Photo &  Pose & Purch. & Sit & SitD. & Smoke & Wait & WalkD. & Walk & WalkT. & Avg.\\
\midrule[.8pt]
Martinez \textit{et al.}~\cite{martinez2017simple} &  51.8 &56.2 &58.1 &59.0 &69.5& 78.4 &55.2 &58.1 &74.0 &94.6 &62.3 &59.1& 65.1& 49.5& 52.4 &62.9\\
Sun \textit{et al.}~\cite{sun2017compositional} &52.8& 54.8& 54.2& 54.3 &61.8 &67.2& 53.1& 53.6 &71.7 &86.7 &61.5 &53.4& 61.6 &47.1& 53.4 &59.1\\
Yang \textit{et al.}~\cite{yang20183d} & 51.5& 58.9&  50.4 &57.0& 62.1& 65.4 &49.8& 52.7& 69.2& 85.2& 57.4& 58.4& 43.6& 60.1& 47.7& 58.6\\
Fang \textit{et al.}~\cite{fang2018learning}& 50.1 &54.3& 57.0& 57.1& 66.6& 73.3& 53.4& 55.7& 72.8& 88.6& 60.3 &57.7& 62.7& 47.5 &50.6& 60.4\\
Hossain \& Little~\cite{rayat2018exploiting}  & 48.4 & 50.7 & 57.2 & 55.2 & 63.1 & 72.6 & 53.0 & 51.7 & 66.1 & 80.9 & 59.0 & 57.3 & 62.4 & 46.6 & 49.6 & 58.3\\
Pavlakos \textit{et al.}~\cite{pavlakos2018ordinal} & 48.5& 54.4& 54.4& 52.0 &59.4 &65.3 &49.9& 52.9& 65.8 &71.1& 56.6& 52.9& 60.9& 44.7& 47.8& 56.2\\
Sharma \textit{et al.}~\cite{sharma2019monocular} & 48.6 &54.5& 54.2& 55.7& 62.2& 72.0& 50.5& 54.3& 70.0& 78.3 &58.1& 55.4& 61.4& 45.2& 49.7& 58.0\\
Zhao \textit{et al.}~\cite{zhao2019semantic} & 47.3& 60.7& 51.4 &60.5& 61.1& \textbf{49.9}& 47.3& 68.1 &86.2& \textbf{55.0}& 67.8& 61.0& \textbf{42.1}& 60.6& 45.3& 57.6\\
Li \textit{et al.}~\cite{ChenLiLee:2020} & 62.0 & 69.7 & 64.3 & 73.6 & 75.1 & 84.8 & 68.7 & 75.0 & 81.2 & 104.3 & 70.2 & 72.0 & 75.0 & 67.0 & 69.0 & 73.9\\
Banik \textit{et al.}~\cite{Banik:2021} & 51.0 & 55.3 & 54.0 & 54.6 & 62.4 & 76.0 & 51.6 & 52.7 & 79.3 & 87.1 & 58.4 & 56.0 & 61.8 & 48.1 & 44.1 & 59.5\\
Xu \textit{et al.}~\cite{YuanluXu:2021} & 47.1 & 52.8 & 54.2 & 54.9 & 63.8 & 72.5 & 51.7 & 54.3 & 70.9 & 85.0 & 58.7 & 54.9 & 59.7 & 43.8 & 47.1 & 58.1\\
Zou \textit{et al.}~\cite{zou2020high} & 49.0& 54.5& 52.3& 53.6& 59.2 &71.6& 49.6& 49.8 &66.0 &75.5 &55.1 &53.8& 58.5& 40.9& 45.4 &55.6\\
Quan \textit{et al.}~\cite{quan2021higher} &47.0 & 53.7 & 50.9 & 52.4&57.8 &71.3&50.2 &49.1 &63.5 &76.3  &54.1&51.6 &56.5 &41.7 &45.3 &54.8 \\
Zou \textit{et al.}~\cite{zou2021compositional} & 48.4 &  53.6 &  49.6 &  53.6 &  57.3 &  70.6 &  51.8 &  50.7 &  62.8 &  74.1 &  54.1 &  52.6 &  58.2 &  41.5 &  45.0 &  54.9\\
Liu \textit{et al.}~\cite{liu2020comprehensive} & 46.3 & 52.2 & 47.3 & 50.7 & 55.5 & 67.1 & 49.2 & 46.0 & 60.4 & 71.1 & 51.5 & 50.1 & 54.5 & 40.3 & 43.7 & 52.4 \\
Lin \textit{et al.}~\cite{Kevin2021End} & - & - & - & - & - & - & - & - & - & - & - & - & - & - & - & 54.0 \\
Gong \textit{et al.}~\cite{Kehong2021PoseAug} & - & - & - & - & - & - & - & - & - & - & - & - & - & - & - & 50.2 \\
Zhao \textit{et al.}~\cite{Weixi2021GraFormer} & 45.2 & 50.8 & 48.0 & 50.0 & 54.9 & 65.0 & 48.2 & 47.1 & 60.2 & 70.0 & 51.6 & 48.7 & 54.1 & 39.7 & 43.1 & 51.8 \\
Zheng \textit{et al.}~\cite{PoseFormer:2021} ($f{=}1$) & 46.9 & 	51.9 & 	46.9 & 	51.2 & 	53.4 & 	60.0 & 	49.0 & 	47.5 & 	58.8 & 	67.2 & 	51.6 & 	48.9 & 	 54.3 & 	40.2 & 	 42.1 & 	51.3 \\
Zhang \textit{et al.}~\cite{Jinlu2022Mix} ($f{=}1$) & 46.0 & 49.9 &  49.1 & 	50.8 & 	52.7 & 	58.4 & 	48.4 & 	47.3 & 	60.3 & 	67.6 & 	51.4 & 	48.5 & 	53.8 & 	39.5 & 	 42.7 & 	 51.1 \\
Zou \textit{et al.}~\cite{zou2021modulated} & 45.4 & 49.2 & 45.7 & \underline{49.4} & 50.4 &  58.2 & 47.9 & 46.0 & 57.5 & 63.0 & 49.7 & 46.6 & 52.2 & \underline{38.9} & \underline{40.8} & 49.4\\
\midrule[.8pt]
Ours (CPN) & \underline{44.7} &	\underline{48.4} &	\underline{44.8} &	49.7 &	\underline{49.6} &	58.2 &	\underline{47.4} &	 \underline{44.8} &	 \underline{55.2} &	 \underline{59.7} &	\underline{49.3} &	\underline{46.4} &	51.4 &	\textbf{38.6} &	\textbf{40.6} &	 \underline{48.6} \\
Ours & \textbf{41.0} &	\textbf{46.8} &	\textbf{44.0} &	\textbf{48.4} &	\textbf{47.5} &	\underline{50.7} &	\textbf{45.4} &	\textbf{42.3} &	 \textbf{53.6} &	 65.8 &	 \textbf{45.6} & \textbf{45.2} & \underline{48.9} & 39.7 & \textbf{40.6} & \textbf{47.0} \\
			
\bottomrule[1pt]
\end{tabular}
\label{Tab:Result1}
\end{table*}

Under Protocol \#2, Table~\ref{Tab:Result2} shows that RS-Net outperforms ModulatedGCN~\cite{zou2021modulated} on 11 out of 15 actions, as well as on average. Our model also performs better than high-order GCN~\cite{zou2020high} with a 11.67\% error reduction on average, achieving better performance on all 15 actions, and indicating the importance of weight and adjacency modulation. Another insight from Tables~\ref{Tab:Result1} and~\ref{Tab:Result2} is that our model outperforms GCN with weight unsharing~\cite{liu2020comprehensive} on all actions under Protocol \#1 and  Protocol \#2, while using a fewer number of learnable parameters. This indicates the usefulness of not only higher-order structural information, but also weight and adjacency modulation in boosting human pose estimation performance.
	
\begin{table*}[!htb]
\caption{Performance comparison of our model and baseline methods using PA-MPJPE on Human3.6M under Protocol \#2.}
\small
\setlength{\tabcolsep}{1pt}
\smallskip
\centering
\begin{tabular}{l*{17}{c}}
\toprule[1pt]
& \multicolumn{15}{c}{Action}\\
\cmidrule(lr){2-16}
Method & Dire. & Disc. &  Eat & Greet & Phone & Photo &  Pose & Purch. & Sit & SitD. & Smoke & Wait & WalkD. & Walk & WalkT. & Avg.\\
\midrule[.8pt]
Zhou \textit{et al.}~\cite{zhou2017towards} & 47.9& 48.8 &52.7& 55.0& 56.8& 49.0 &45.5 &60.8& 81.1 &53.7& 65.5& 51.6& 50.4 &54.8 &55.9& 55.3\\
Pavlakos \textit{et al.}~\cite{pavlakos2017coarse} & 47.5 &50.5 &48.3& 49.3& 50.7 &55.2 &46.1 &48.0& 61.1& 78.1 &51.1& 48.3& 52.9& 41.5& 46.4 &51.9 \\
Martinez \textit{et al.}~\cite{martinez2017simple} & 39.5 &43.2 &46.4 &47.0 &51.0& 56.0 &41.4& 40.6 &56.5& 69.4& 49.2& 45.0& 49.5& 38.0 &43.1 &47.7\\
Sun \textit{et al.}~\cite{sun2017compositional} & 42.1& 44.3& 45.0 &45.4 &51.5 &53.0 &43.2& 41.3& 59.3 &73.3& 51.0& 44.0& 48.0& 38.3& 44.8& 48.3\\
Fang \textit{et al.}~\cite{fang2018learning} & 38.2& 41.7& 43.7& 44.9& 48.5 &55.3& 40.2& 38.2& 54.5 &64.4& 47.2 &44.3& 47.3& 36.7& 41.7& 45.7\\
Hossain \& Little~\cite{rayat2018exploiting}  & 35.7 & 39.3& 44.6 &43.0& 47.2& 54.0& 38.3 &37.5 &51.6 &61.3& 46.5& 41.4 &47.3 &34.2 &39.4& 44.1\\
Li \textit{et al.}~\cite{ChenLiLee:2020} & 38.5 & 41.7 & 39.6 & 45.2 & 45.8 & 46.5 & 37.8 & 42.7 & 52.4 & 62.9 & 45.3 & 40.9 & 45.3 & 38.6 & 38.4 & 44.3\\
Banik \textit{et al.}~\cite{Banik:2021} & 38.4 & 43.1 & 42.9 & 44.0 & 47.8 & 56.0 & 39.3 & 39.8 & 61.8 & 67.1 & 46.1 & 43.4 & 48.4 & 40.7 & 35.1 & 46.4\\
Xu \textit{et al.}~\cite{YuanluXu:2021} & 36.7 & 39.5 & 41.5 & 42.6 & 46.9 & 53.5 & 38.2 & 36.5 & 52.1 & 61.5 & 45.0 & 42.7 & 45.2 & 35.3 & 40.2 & 43.8\\
Zou \textit{et al.}~\cite{zou2020high} &38.6 &42.8& 41.8 &43.4 &44.6& 52.9& 37.5& 38.6 & 53.3 & 60.0 & 44.4 & 40.9 & 46.9 & 32.2 &37.9 & 43.7\\
Quan \textit{et al.}~\cite{quan2021higher} & 36.9 & 42.1&40.3 &42.1 &43.7 &52.7&37.9 &37.7 &51.5 &60.3  &43.9&39.4 & 45.4 & 31.9 & 37.8 & 42.9 \\
Zou \textit{et al.}~\cite{zou2021compositional} & 38.4  & 41.1 &  40.6 &  42.8 &  43.5 &  51.6 &  39.5 &  37.6 &  49.7 &  58.1 &  43.2 &  39.2 &  45.2 &  32.8 &  38.1 &  42.8\\
Liu  \textit{et al.}~\cite{liu2020comprehensive} & 35.9 & 40.0 & 38.0 & 41.5 & 42.5 & 51.4 & 37.8 & 36.0 & 48.6 & 56.6 & 41.8 & 38.3 & 42.7 & 31.7 & 36.2 & 41.2\\
Zheng \textit{et al.}~\cite{PoseFormer:2021} ($f{=}1$) & 36.0 & 39.5 & 	37.4 & 	40.9 & 	40.5 & 	45.6 & 	\underline{36.4} & 	35.6 & 	47.9 & 	53.9 & 	41.4 & 	 36.5 & 	 42.3 & 	 30.8 & 	34.3 & 	39.9 \\
Zhang \textit{et al.}~\cite{Jinlu2022Mix} ($f{=}1$) & 36.1	& 38.9 & 38.8 & 41.1 & 	40.2 & 	45.0 & 	37.2 & 	36.2 & 	48.9 & 	54.1 & 	41.1 & 	36.7 & 	42.4 & 	 31.1 & 	 35.2 & 	 40.2 \\
Zou \textit{et al.}~\cite{zou2021modulated} & 35.7 & 38.6 & 36.3 & \textbf{40.5} & \textbf{39.2} & \underline{44.5} & 37.0 & 35.4 & 46.4 &  \underline{51.2} & 40.5 & \underline{35.6} & 41.7 & \textbf{30.7} & \underline{33.9} & 39.1\\
\midrule[.8pt]
Ours (CPN) & \underline{35.5} &	\underline{38.3} &	\underline{36.1} &	\textbf{40.5} &	\textbf{39.2} &	44.8 &	37.1 &	 \underline{34.9} &	\underline{45.0} &	 \textbf{49.1} &	 \underline{40.2} &	\textbf{35.4} &	\underline{41.5} &	\underline{31.0} &	34.3 &	 \underline{38.9} \\
Ours & \textbf{34.2} &	\textbf{38.2} &	\textbf{35.6} &	\underline{40.8} &	\underline{38.5} &	\textbf{41.8} &	\textbf{36.0} &	\textbf{34.0} &	 \textbf{43.9} & 56.2 &	 \textbf{38.0}	& 36.3 & \textbf{40.2} & 31.2 & \textbf{33.3} & \textbf{38.6}  \\
\bottomrule[1pt]
\end{tabular}
\label{Tab:Result2}
\end{table*}

In Table~\ref{Tab:MPI}, we report the quantitative comparison results of RS-Net and several baselines on the MPI-INF-3DHP dataset. As can be seen, our method achieves significant improvements over the comparative methods. In particular, our model outperforms the best baseline (i.e., PoseFormer) with relative improvements of 1.42\% and 2.11\% in terms of the PCK and AUC metrics, respectively. Overall, our model consistently outperforms the baseline methods in terms of all evaluation metrics on both datasets, indicating its effectiveness in 3D human pose estimation.

\begin{table}[!htb]
\caption{Performance comparison of our model and baseline methods on the MPI-INF-3DHP dataset using PCK and AUC as evaluation metrics. Higher values in boldface indicate the best performance, and the best baselines are underlined.}
\small
\setlength{\tabcolsep}{2.5pt}
\medskip
\centering
\begin{tabular}{lcc}
\toprule
Method & PCK($\uparrow$) & AUC($\uparrow$)\\
\midrule
Chen  \textit{et al.}~\cite{li2019generating} & 67.9 & - \\
Yang \textit{et al.}~\cite{yang20183d} & 69.0 & 32.0 \\
Pavlakos \textit{et al.}~\cite{pavlakos2018ordinal}  & 71.9 & 35.3 \\
Habibie \textit{et al.}~\cite{Habibie:19}  & 70.4 & 36.0 \\
Quan \textit{et al.}~\cite{quan2021higher} & 72.8 &36.5 \\
Zhao \textit{et al.}~\cite{Weixi2021GraFormer} & 79.0 & 43.8\\
Zeng \textit{et al.}~\cite{Ailing2021Hard} & 82.1 & 46.2 \\
Zou \textit{et al.}~\cite{zou2021compositional} & 79.3 & 45.9\\
Zheng \textit{et al.}~\cite{PoseFormer:2021} ($f{=}1$) & \underline{84.4} & \underline{52.1} \\
\midrule
Ours & \textbf{85.6} &\textbf{53.2} \\
\bottomrule
\end{tabular}
\label{Tab:MPI}
\end{table}
	
\begin{figure*}[!htb]
\centering
\includegraphics[scale=.43,angle=-90]{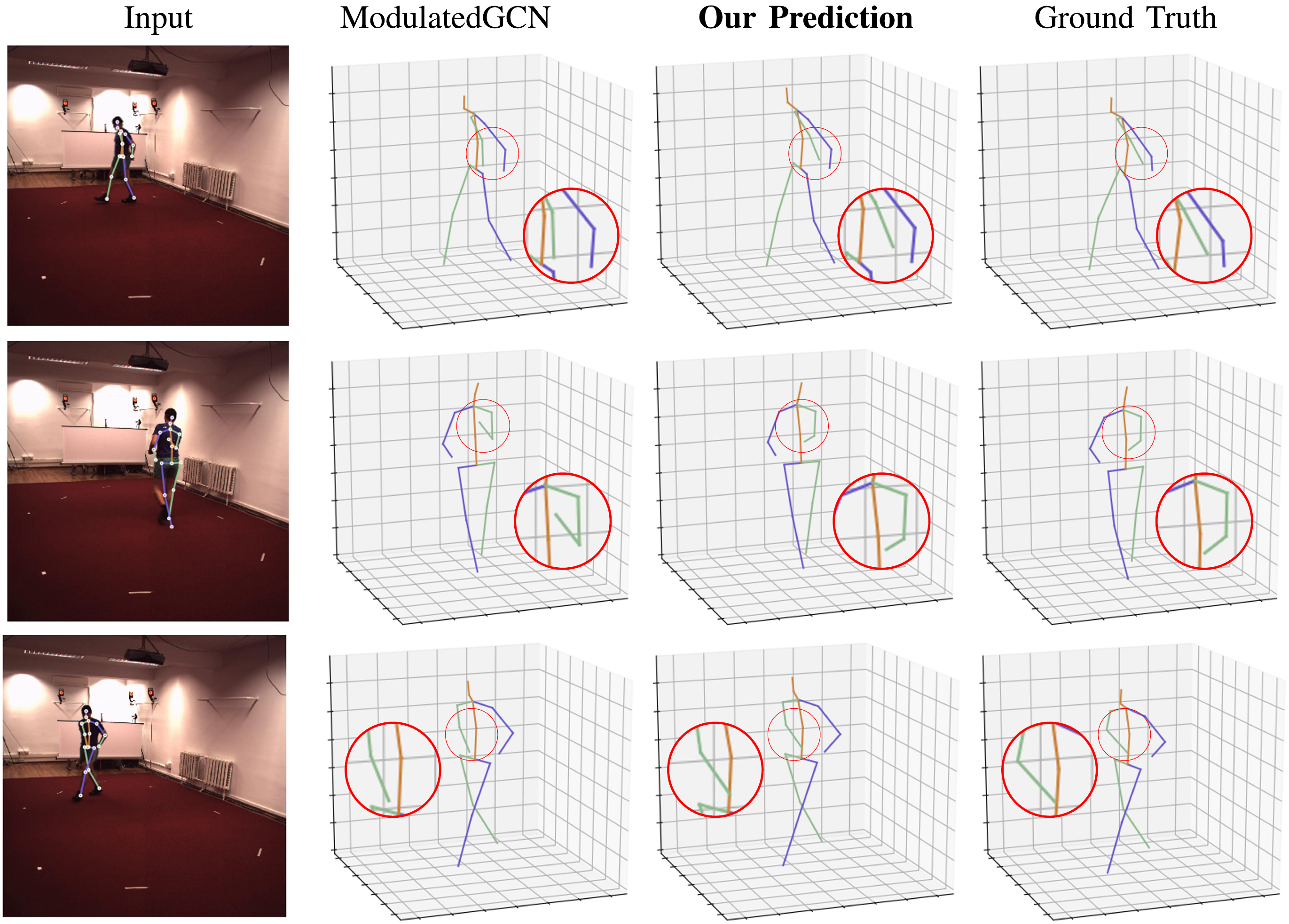}
\caption{Qualitative comparison between our model and ModulatedGCN on the Human3.6M dataset for different actions. The red circle indicates the locations where our model yields better results.}
\label{Fig:visual}
\end{figure*}

\medskip\noindent\textbf{Qualitative Results.}\quad Figure~\ref{Fig:visual} shows the qualitative results obtained by the proposed RS-Net model for various actions. As can be seen, the predictions made by our model are better than ModulatedGCN and match more closely the ground truth, indicating the effectiveness of RS-Net in tackling the 2D-to-3D human pose estimation problem. Notice that ModulatedGCN fails to properly predict the hand poses when there are occlusions. In comparison, our model is able to reliably predict the hand poses.
	
\subsection{Ablation Studies}
In order to verify the impact of the various components on the effectiveness of the proposed RS-Net model, we conduct ablation experiments on the Human3.6M dataset under Protocol \#1 using MPJPE as evaluation metric.

\medskip\noindent\textbf{Effect of Skip Connection.}\quad We start by investigating the impact of the initial skip connection on model performance. Results reported in Table~\ref{Tab:initialSkipConnection} show that skip connection helps improve the performance of our model, yielding relative error reductions of .58\% and .74\% in terms of MPJPE and PA-MPJPE, respectively. While these improvements may not sound significant, they, however, add up because the evaluation metrics are measured in millimeters.

\begin{table}[!htb]
\caption{Effectiveness of initial skip connection (ISC). Boldface numbers indicate the best performance.}
\small
\setlength{\tabcolsep}{2.5pt}
\medskip
\centering
\begin{tabular}{l*{7}{c}}
\toprule
Method & Filters  & Param. & MPJPE($\downarrow$) & PA-MPJPE($\downarrow$)\\
\midrule
w/o ISC & 64 & 0.7M & 51.7 & 40.4\\
w/ ISC & 48 & 0.7M & \textbf{51.4} & \textbf{40.1} \\
\bottomrule
\end{tabular}
\label{Tab:initialSkipConnection}
\end{table}
	
\begin{figure}[!htb]
\centering
\setlength{\tabcolsep}{1pt}
\begin{tabular}{cc}
\includegraphics[width=1.75in]{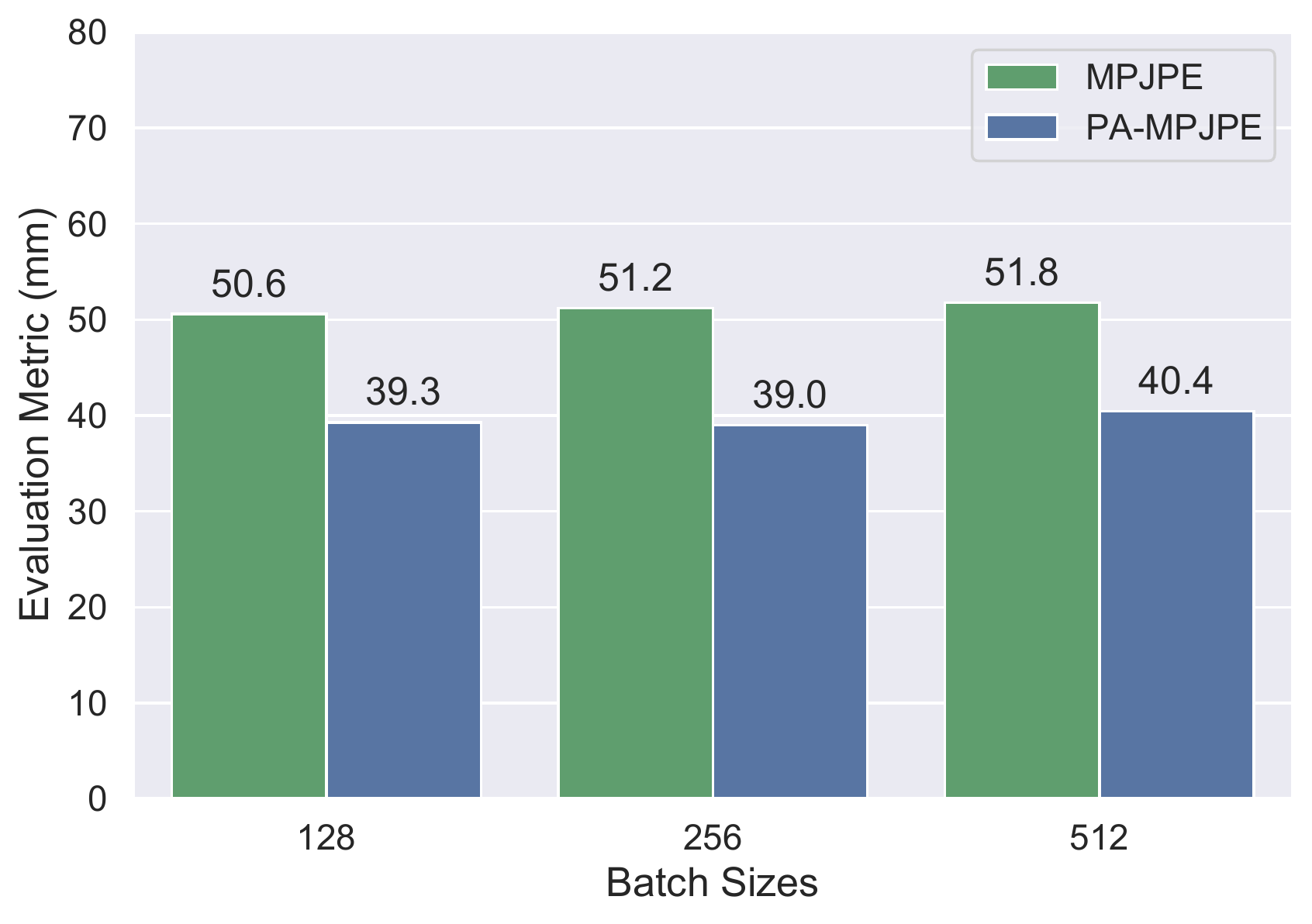} & \includegraphics[width=1.75in]{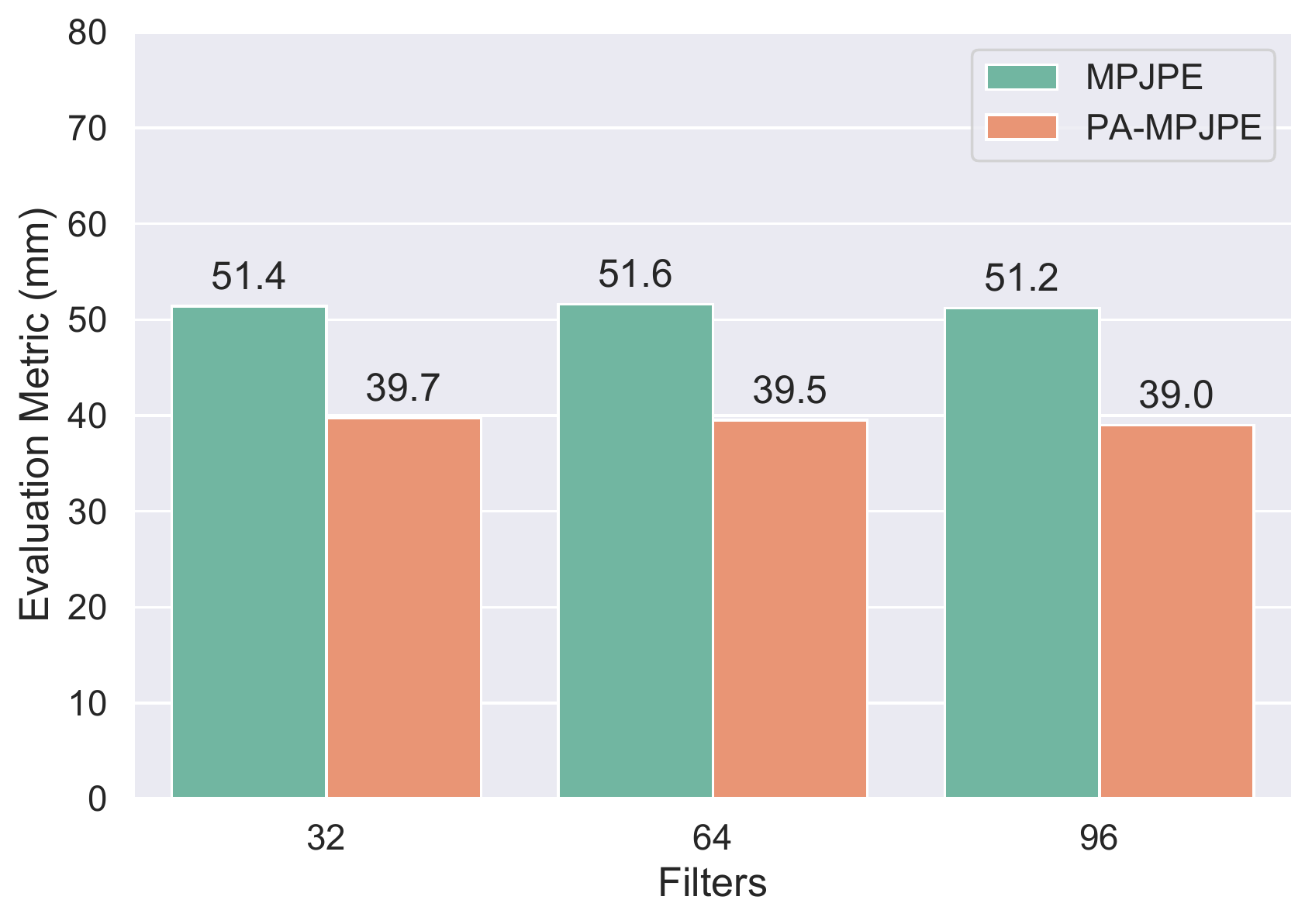}
\end{tabular}
\caption{Performance of our proposed RS-Net model on the Human3.6M dataset using various batch and filter sizes.}
\label{Fig:BatchAndFilter}
\end{figure}

\medskip\noindent\textbf{Effect of Batch/Filter Size.}	\quad We also investigate the effect of using different batch and filter sizes on the performance of our model. We report the results in Figure~\ref{Fig:BatchAndFilter}, which shows that the best performance is achieved using a batch size of 128. Similarly, filter sizes of 96 and 64 yield the best performance in terms of MPJPE and PA-MPJPE, respectively.

\begin{figure}[!htb]
\centering
\setlength{\tabcolsep}{1pt}
\begin{tabular}{cc}
\includegraphics[width=1.75in]{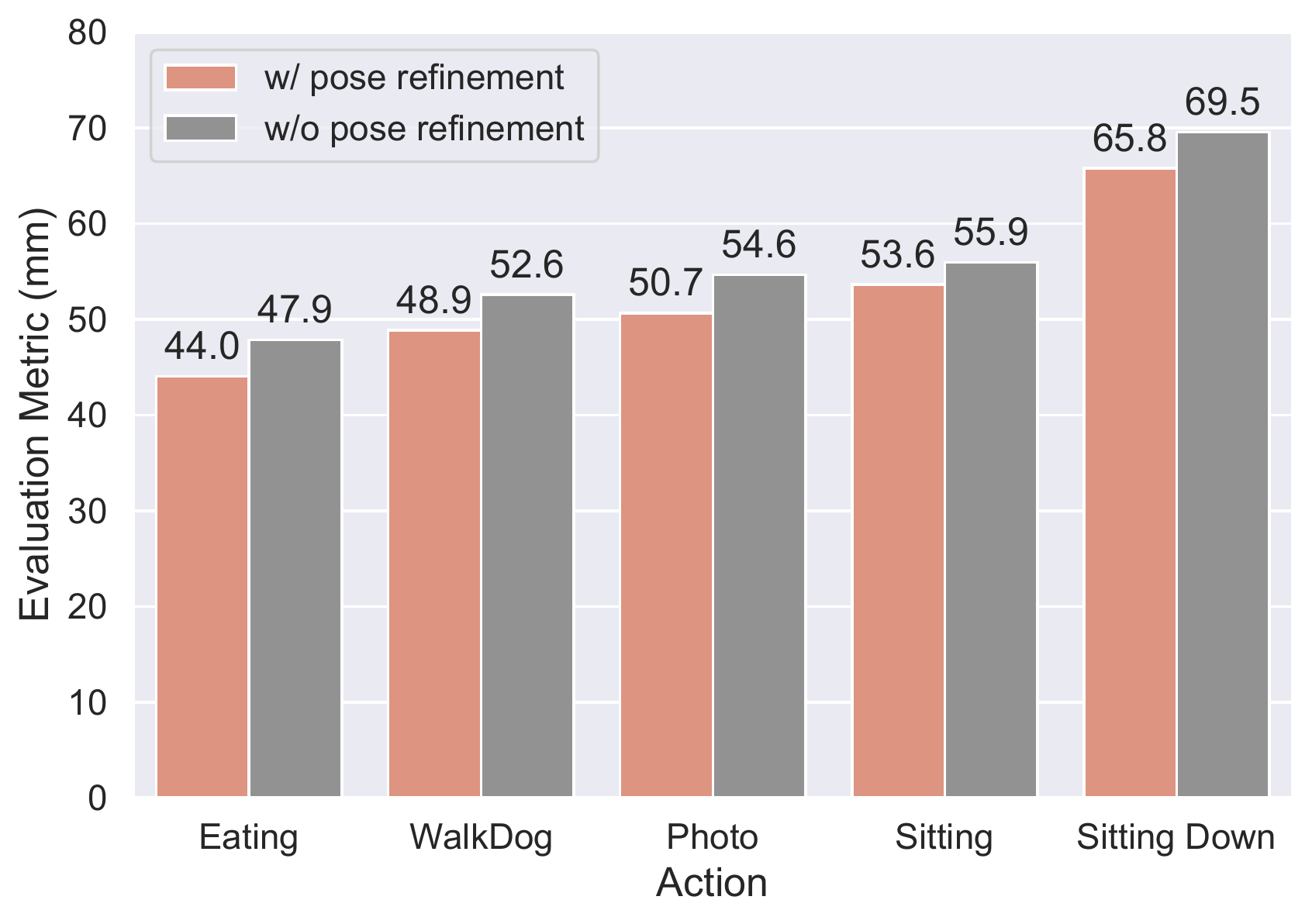} & \includegraphics[width=1.75in]{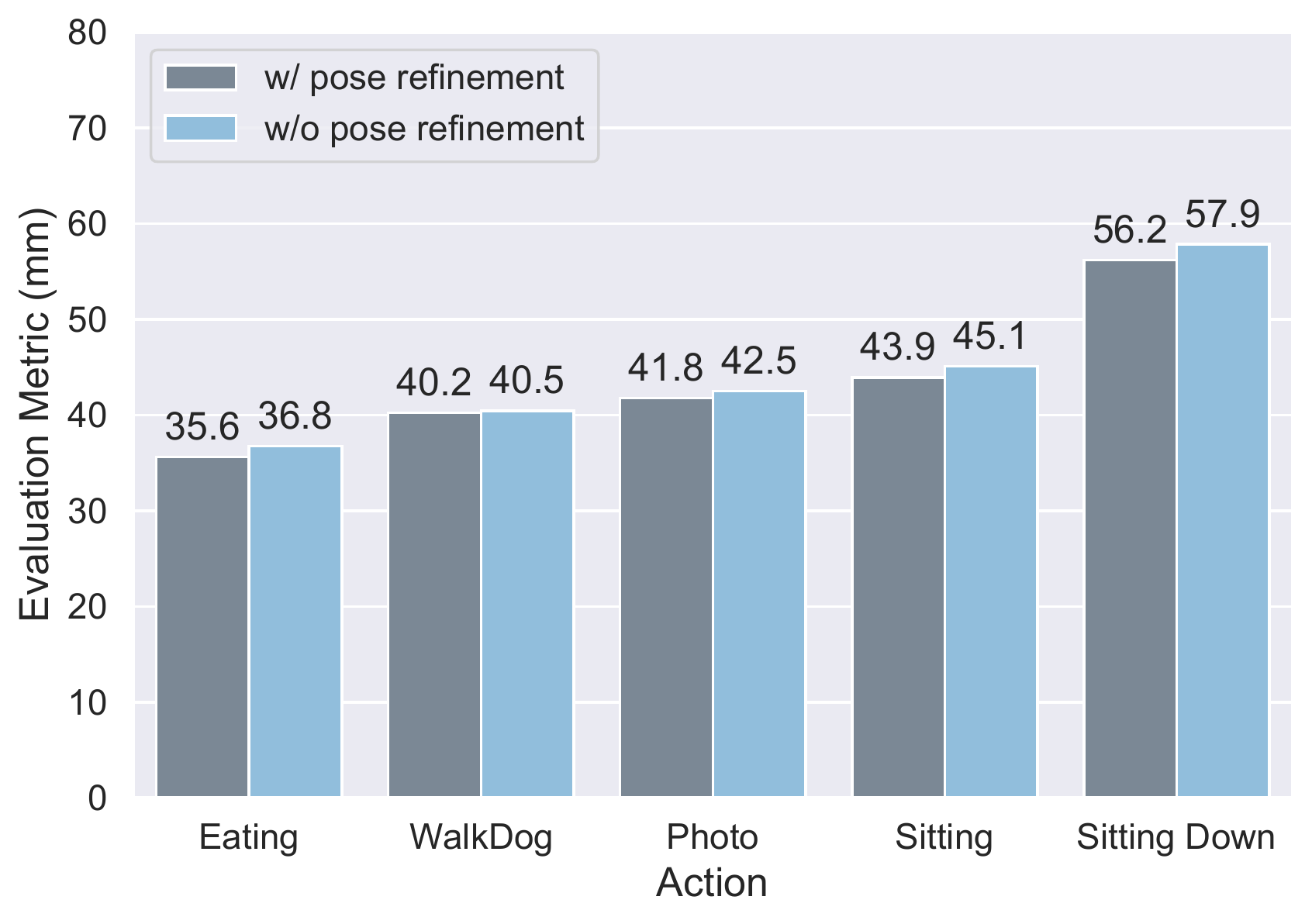}
\end{tabular}
\caption{Performance of our model with and without pose refinement using MPJPE (top) and PA-MPJPE (bottom).}
\label{Fig:post_refine}
\end{figure}

\medskip\noindent\textbf{Effect of Pose Refinement.}\quad Following~\cite{YujunCai:19}, we use a pose refinement network, which is comprised of two fully connected layers. Pose refinement helps improve the estimation accuracy of 3D joint locations. Through experimentation, we find that using a batch size of 512 with pose refinement yields improvements around .52 mm in MPJPE and .32 mm in PA-MPJPE compared to a batch size of 128. Figure~\ref{Fig:post_refine} shows the performance of our model with and without pose refinement under Protocol \#1 (left) and Protocol \#2 (right). As can be seen, lower errors are obtained when integrating pose refinement into our model, particularly under Protocol \#1 for various human actions. In the case of the ``Sitting Down'' action, for example, pose refinement yields an error reduction of 5.32\% in terms of MPJPE.
	
\medskip\noindent\textbf{Effect of Residual Block Design.}\quad In Table~\ref{Tab:LayerNormAndGELU}, we report the comparison results between two residual block designs: the first design employs blocks consisting of convolutional layers followed by batch normalization (BatchNorm) and a ReLU activation function, while the second design uses blocks comprised of convolutional layers followed by layer normalization (LayerNorm) and a GELU activation function, which is a smoother version of ReLU and is commonly used in Transformers based approaches. As can be seen, using the ConvNext architectural block design, we obtain relative performance gains of 1.67\% and 1.28\% in terms of MPJPE and PA-MPJPE, respectively.

\begin{table}[!htb]
\caption{Effect of residual block design of the performance of our model. We use filters of size 96. Lower values in boldface indicate the best performance.}
\small
\setlength{\tabcolsep}{2.5pt}
\medskip
\centering
\begin{tabular}{l*{7}{c}}
\toprule
Method  & MPJPE($\downarrow$) & PA-MPJPE($\downarrow$)\\
\midrule
Ours w/ BatchNorm and ReLU & 47.8 & 39.1 \\
Ours w/ LayerNorm and GELU   & \textbf{47.0} & \textbf{38.6}  \\
\bottomrule			
\end{tabular}
\label{Tab:LayerNormAndGELU}
\end{table}

We also compare our model to ModulatedGCN~\cite{zou2021modulated}, Weight Unsharing~\cite{liu2020comprehensive}, SemGCN~\cite{zhao2019semantic}, and High-order GCN~\cite{zou2020high} using ground truth keypoints, and we report the results in Table~\ref{Tab:baseline}. As can be seen, our model consistently performs better than these baselines under both Protocols \#1 and \#2. Under Protocol \#1, our RS-Net model outperforms ModulatedGCN, Weight Unsharing, High-order GCN and SemGCN by .15 mm, .55 mm, 2.24 mm and 3.50 mm, which correspond to relative error reductions of .40\%, 1.45\%, 5.67\%, and 8.58\%, respectively. Under Protocol \#2, our RS-Net model performs better than ModulatedGCN, Weight Unsharing, High-order GCN, and SemGCN by .66 mm, 1.02 mm, 2 mm and 2.39 mm, which translate into relative improvements of 2.22\%, 3.39\%, 6.44\% and 7.60\%, respectively.

\begin{table}[!htb]
\caption{Performance comparison of our model and other GCN-based methods without pose refinement using ground truth keypoints. Boldface numbers indicate the best performance.}
\small
\setlength{\tabcolsep}{2.5pt}
\medskip
\centering
\begin{tabular}{l*{7}{c}}
\toprule
Method & Filters & Param. & MPJPE($\downarrow$) & PA-MPJPE($\downarrow$)\\
\midrule
SemGCN~\cite{zhao2019semantic} & 128 & 0.43M & 40.78 & 31.46 \\
High-order GCN~\cite{zou2020high}  & 96 & 1.20M & 39.52 &31.07 \\
Weight Unsharing~\cite{liu2020comprehensive} & 128 & 4.22M & 37.83 & 30.09  \\
ModulatedGCN~\cite{zou2021modulated} & 256 & 1.10M & 37.43 & 29.73 \\
\midrule
Ours & 64 & 1.77M & \textbf{37.28} & \textbf{29.07} \\
	
\bottomrule
\end{tabular}
\label{Tab:baseline}
\end{table}

\begin{figure}[!htb]
\centering
\includegraphics[width=3.3in]{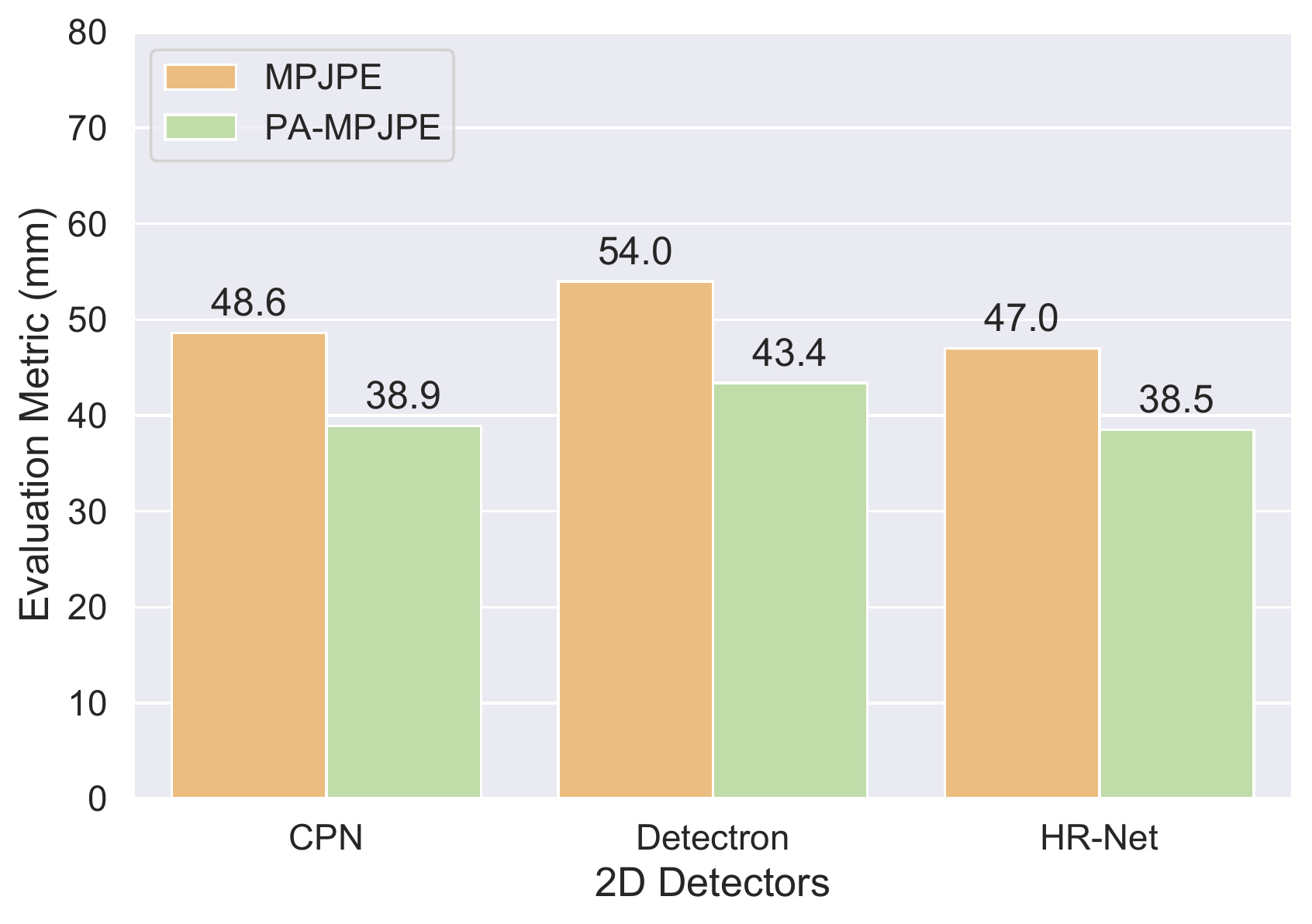}
\caption{Performance of our model with pose refinement using different 2D detectors.}
\label{Fig:detectors}
\end{figure}

In order to gain further insight into the importance of pose refinement, we train our model with pose refinement on the Human3.6M dataset using 2D poses from three different 2D pose detectors, including cascaded pyramid network (CPN)~\cite{chen2018cascaded}, Detectron~\cite{wu2019detectron2} and high-resolution network (HR-Net)~\cite{sun2019deep}. As shown in Figure~\ref{Fig:detectors}, the best performance is achieved using the HR-Net detector in terms of both MPJPE and PA-MPJPE.
	
\section{Conclusion}
In this paper, we introduced an effective higher-order graph network with initial skip connection for 3D human pose estimation using regular matrix splitting in conjunction with weight and adjacency modulation. The aim is to capture not only the long-range dependencies between body joints, but also the different relations between neighboring joints and distant ones. In our proposed model architecture, we designed a variant of the ConvNeXt residual block, comprised of convolutional layers, followed by layer normalization and a GELU activation function. Experimental results on two standard benchmark datasets demonstrate that our model can outperform qualitatively and quantitatively several recent state-of-the-art methods for 3D human pose estimation. For future work, we plan to incorporate temporal information into our model by constructing a spatiotemporal graph on skeleton sequences and exploiting both spatial and temporal relationships between body joints in order to further improve the 3D pose estimation accuracy.

\section*{Acknowledgements}
This work was supported in part by the Natural Sciences and Engineering Research Council of Canada (NSERC).

\bibliographystyle{ieeetr}
\bibliography{references} 

\end{document}